\documentclass[final]{cvpr}

\usepackage{times}
\usepackage{epsfig}
\usepackage{graphicx}
\usepackage{amsmath}
\usepackage{amssymb}

\usepackage{array}
\newcolumntype{C}[1]{>{\centering\let\newline\\\arraybackslash\hspace{0pt}}m{#1}}
\usepackage{algorithm}
\usepackage[noend]{algpseudocode}
\usepackage{algorithmicx,algorithm}
\usepackage{multirow}
\usepackage{xcolor}
\usepackage{color, colortbl}
\usepackage{hhline}
\usepackage{makecell}
\usepackage{subfigure}
\definecolor{Gray}{gray}{0.9}

\usepackage{array}
\newcolumntype{C}[1]{>{\centering\arraybackslash}p{#1}}

\newcommand{\Cov}{\mathrm{Cov}}
\usepackage[pagebackref=true,breaklinks=true,colorlinks,bookmarks=false]{hyperref}

\def\cvprPaperID{4970} % *** Enter the CVPR Paper ID here
\def\confYear{CVPR 2021}

\begin{document}

%%%%%%%%% TITLE
\title{Convolutional Neural Network Pruning with Structural Redundancy Reduction}

\author{Zi Wang$^1$, Chengcheng Li$^1$, Xiangyang Wang$^2$\\
$^1$The University of Tennessee, Knoxville, TN, USA\\
$^2$Sun Yat-sen University, Guangzhou, China\\
{\tt\small \{zwang84,cli42\}@vols.utk.edu, mcswxy@mail.sysu.edu.cn}
% \and
}

\maketitle
% \pagestyle{empty}
% \thispagestyle{empty}

%%%%%%%%% ABSTRACT
\begin{abstract}
Convolutional neural network (CNN) pruning has become one of the most successful network compression approaches in recent years. Existing works on network pruning usually focus on removing the least important filters in the network to achieve compact architectures. In this study, we claim that identifying structural redundancy plays a more essential role than finding unimportant filters, theoretically and empirically. We first statistically model the network pruning problem in a redundancy reduction perspective and find that pruning in the layer(s) with the most structural redundancy outperforms pruning the least important filters across all layers. Based on this finding, we then propose a network pruning approach that identifies structural redundancy of a CNN and prunes filters in the selected layer(s) with the most redundancy. Experiments on various benchmark network architectures and datasets show that our proposed approach significantly outperforms the previous state-of-the-art.
\end{abstract}

\section{Introduction}
\label{sec:intro}
Convolutional neural networks (CNNs) \cite{lecun2015deep} have developed substantially in recent years and are widely used in various applications, such as object classification \cite{deng2009imagenet,krizhevsky2012imagenet}, image synthesis, \cite{goodfellow2014generative,tulyakov2018mocogan}, super-resolution \cite{dong2015image}, and game-playing \cite{mnih2015human,silver2016mastering}. State-or-the-art performance are achieved by designing wider and deeper CNNs \cite{simonyan2014very,he2016deep,huang2017densely}. However, the over-parameterization problem of CNNs prevents them from being applied to resource-limited devices, such as mobile phones and robotics \cite{sandler2018mobilenetv2,ma2018shufflenet}. Many approaches have been proposed to reduce the computation and storage cost of CNNs, such as quantization \cite{han2015deep}, matrix decomposition \cite{yu2017compressing}, network pruning \cite{han2015learning,li2016pruning,yu2018slimmable,wang2019towards,he2019filter}, and knowledge distillation \cite{hinton2015distilling}. Network pruning is one of the most popular methods and attracts enormous attention.

Generally, network pruning can be categorized into weight (unstructured) pruning \cite{han2015learning} and channel (structured) pruning \cite{li2016pruning,molchanov2016pruning,wang2019eigendamage,yu2018slimmable}. Weight pruning zeros out specific weights in filters and results in unstructured sparsities. To accelerate the pruned CNNs, specialized hardware and software have to be developed \cite{han2016eie}. Channel pruning, which removes the whole convolutional filters, is a more flexible method without the need for special hardware. 
As the entire filters are deleted, a considerable pruning ratio can usually be achieved with little performance degradation. Many of the existing channel pruning approaches rely on finding and pruning the least important filters, or the filters that share the most similarities with others across all layers \cite{li2016pruning,molchanov2016pruning,he2019filter,ding2019approximated}. For example, \cite{molchanov2016pruning} uses the Taylor series to estimate the loss change after each filter's removal and prune the filters that cause minimal training loss change. It has been a common belief that with a better filter ranking criterion, there is a better chance to drop the least important filters and get a compact network with less performance loss.
% cluster based method find the the filters that share most similarities with others, which is functionally replaceable and can be removed. {\color{red}use a toy case to demonstrate that this is not intuitively true in some cases? like the Geometric Median CVPR paper.}

However, our studies on channel pruning contradict this common belief. Using statistical modeling to measure the redundancy in each convolutional layer, we theoretically show that (in certain cases, even randomly) pruning filters in the layer with the most redundancy outperforms pruning the least important filters across all layers. To our best knowledge, this is the first study that theoretically analyzes the rationale behind network pruning from a redundancy reduction perspective. With this finding, we propose a layer-adaptive channel pruning approach based on structural redundancy reduction (SRR), which is achieved by establishing a graph for each convolutional layer of a CNN and using two quantities associated with the graph, i.e., $\ell$-covering number and quotient space size, as the measurement of the redundancy in each layer. After that, unimportant filters in the identified layer(s) with the most redundancy, rather than the least important filters across all layers, are pruned. 

We summarize the contribution of this study as follows. (1) We theoretically analyze network pruning with statistical modeling from a perspective of redundancy reduction. We find that pruning in the layer(s) with the most redundancy outperforms pruning the least important filters across all layers.
(2) We propose a layer-adaptive channel pruning approach based on structural redundancy reduction, which builds a graph for each convolutional layer of a CNN to measure the redundancy existed in each layer. This approach prunes unimportant filters in the most redundant layer(s), rather than the filters with the least importance across all layers. (3) We validate the proposed approach on various network architectures and datasets. Experiment results demonstrate that our approach achieves state-of-the-art performance compared with recent channel pruning methods. More specifically, our pruned ResNet50 model on ImageNet can reduce 44.1\% FLOPs while losing only $0.37\%$ top-1 accuracy.

\section{Related work}
\label{sec:work}
% \noindent{\bf Early works and weight pruning.}
\subsection{Early works and weight pruning}
Network pruning is a long-standing topic that can be traced back to the 1990s \cite{hanson1989comparing,lecun1990optimal}. In the era of deep learning, \cite{han2015learning} is one of the most famous early works that prunes weights below a threshold. After that, various weight pruning approaches have been proposed \cite{chen2015compressing,liu2018efficient,zhang2018systematic}. As mentioned before, weight pruning causes unstructured sparsities in a network, which is difficult to be used without specialized software and hardware \cite{han2016eie}. 

\subsection{Channel pruning}
% \noindent{\bf Channel pruning.} 
Channel pruning \cite{li2016pruning,molchanov2016pruning,he2019filter,peng2019collaborative} removes the entire filters in a network so that there is no need for specialized hardware. Among all channel pruning approaches, identifying and pruning the least important filters is one of the most popular branches, and can be further divided into three categories. (1) Ranking and pruning filters with a certain criterion. \cite{li2016pruning} and \cite{polyak2015channel} prune the filters with small weight magnitudes or activation values in the corresponding feature maps. \cite{hu2016network} uses the average percentage of zero (APoZ) activation neurons as the criterion and deletes the filters with small ApoZ. First and second-order Taylor expansion are used to estimate the loss change after each filter's removal and the filters that cause minimal loss change are removed \cite{molchanov2016pruning,zeng2018mlprune}. HRank \cite{lin2020hrank} leverages the information in the feature maps to rank the filters. (2) Reconstruction error minimization. Thinet \cite{luo2017thinet} and NISP \cite{yu2018nisp} prune the filters whose removal leads to minimal reconstruction error of the next layer. (3) Similarity measurement. These approaches use various strategies, such as geometric median \cite{he2019filter} and clustering \cite{zhou2018online,duggal2019cup}, to identify the most replaceable filters, or those functionally share the most similarity with others.

\subsection{Pruning as network structure optimization}
% \noindent{\bf Pruning as network structure optimization.} 
Recently, a number of empirical studies indicate that the network structure after pruning, rather than the removal of unimportant filters, plays a decisive role in maintaining the performance of a network. \cite{liu2018rethinking} trained several compact networks obtained by pruning approaches but with random initialization. Surprisingly, comparable or even better performance can be achieved compared with fine-tuning the pruned models. \cite{mittal2018recovering} reports that a network's performance can be recovered even after random pruning. Related to these works, we also find that pruning unimportant filters is not always essential. But beyond that, we theoretically show that pruning in the layers with large redundancy outperforms pruning the least important filters and propose to prune a network based on structural redundancy reduction. 
% Similar to our work, principal component analysis (PCA) is used in previous works to identify the redundancy in CNNs \cite{suau2018network,garg2018single}. But in contrast, our study is with theoretical analysis and shows improvements over the PCA-based approach (see Section \ref{sec:exp}).

% a small part of related work by adding regularization, for example, ADMM \cite{zhang2018systematic}, yihui he's paper (LASSO regression) \cite{he2017channel}. out of scope?
\section{A theoretic analysis of network pruning}
\label{sec:analysis}
We statistically formulate the channel pruning problem from a redundancy reduction perspective. In our context, layer redundancy refers to the number of filters in a convolutional layer. We will later show that the redundancy can be measured with other quantities in real applications. Suppose we have a two-layer CNN\footnote{This configuration can be extended to a multi-layer network (number of layers $\ge 3$) with no difficulty.} with $m$ and $n$ filters, where $n \gg m$. Let $\{\xi_1,\xi_2,\cdots,\xi_m\}$ and $\{\eta_1,\eta_2,\cdots,\eta_n\}$ be one dimensional positive random variables (RVs) representing each filter's contribution to the network performance. For example, a filter's contribution can be represented as the absolute value of $\text{\emph{training accuracy drop}}$ or $\text{\emph{training loss change}}$ after pruning that filter. We call the two layers $\xi$ layer and $\eta$ layer for convenience. We first highlight our finding and then prove it from a statistical modeling perspective.

{\bf Claim:} If a layer has much higher redundancy, pruning filters in that layer, either randomly or selectively, outperforms pruning the least important filters across all layers. %\note{}

%$\zeta = min\{\xi_1,...,\xi_m,\eta_1,...\eta_n\}$

We choose positive constants $a, b >0$, and use the random events $(\sum_{i=1}^m \xi_i \ge a)$ and $(\sum_{i=1}^n \eta_i \ge b)$ to describe the layers $\xi$ and $\eta$ ``performing well''. Then the performance of a system (i.e., the whole neural network) $p$ is measured by the sum of probabilities of the two events (see Equation \eqref{eq:vo}). We define one system ($p_1$) to perform better than another ($p_2$) if $p_1 > p_2$. A natural question is, if we prune a filter from the network, i.e., remove one variable from $\{\xi_1, \xi_2, \cdots, \xi_m, \eta_1,\eta_2,\cdots,\eta_n\}$, how does the system performance change? There are the following cases (the performances of the systems are listed in Equations \eqref{eq:vo}-\eqref{eq:gl}):  (1) no pruning; (2) randomly pruning a filter in the $\eta$ layer, without loss of generality, we assume the last one $\eta_n$ is pruned;  (3) pruning the least important filter $\underline{\eta}=\min\{\eta_1,...,\eta_n\}$ in the $\eta$ layer; (4) pruning the least important filter $\underline{\xi}=\min\{\xi_1,...,\xi_m\}$ in the $\xi$ layer; and (5) pruning the globally least important filter, i.e.,  $\min\{\underline{\xi},\underline{\eta}\}$.
\small
% \vskip -0.1in
\begin{equation}
p_o = P(\sum_{i=1}^m\xi_i \ge a) + P(\sum_{i=1}^n\eta_i \ge b)
\label{eq:vo}
\end{equation}
% \vskip -0.13in
\begin{equation}
p_{\eta r} = P(\sum_{i=1}^m\xi_i \ge a) + P(\sum_{i=1}^{n-1}\eta_i \ge b) %\footnote{Since all variables are \emph{i.i.d.}, we can consider that $\eta_n$ is pruned for simplicity.}
\label{eq:sr}
\end{equation}
% \vskip -0.12in
\begin{equation}
p_{\underline{\eta}} = P(\sum_{i=1}^m\xi_i \ge a) + P(\sum_{i=1}^{n}\eta_i - \underline{\eta} \ge b) \label{eq:sl}
\end{equation}
% \vskip -0.12in
\begin{equation}
p_{\underline{\xi}} = P(\sum_{i=1}^m\xi_i - \underline{\xi} \ge a) + P(\sum_{i=1}^{n}\eta_i \ge b)
\label{eq:fl}
\end{equation}
% \vskip -0.1in
\begin{equation}
\begin{aligned}
p_{g} = \frac{m}{m+n} p_{\underline{\xi}} +\frac{n}{m+n} p_{\underline{\eta}}
\end{aligned}
\label{eq:gl}
\end{equation}
\normalsize
It is worth mentioning that we consider the network performance from a perspective of redundancy (or capacity). That is why we do not divide the probabilities in Equations. \eqref{eq:vo}-\eqref{eq:gl} by $m$ or $n$. a or b can be considered as a threshold. As long as the total contribution of the filters in a layer is greater than the threshold, there is no performance loss. If a layer has too much redundancy (too many filters in our context), then it’s very likely that the total contribution of the filters can still be greater than the threshold after pruning some of them.

Note that $0 \le \eta_n - \underline{\eta} \le \eta_n$, we have
\small
% \vskip -0.17in
% \begin{equation*}
% (\sum_{i=1}^{n-1}\eta_i \ge b) \subset (\sum_{i=1}^{n}\eta_i - \underline{\eta} \ge b) \subset (\sum_{i=1}^{n}\eta_i \ge b),
% \end{equation*}
% \vskip -0.15in
\begin{equation}
P(\sum_{i=1}^{n-1}\eta_i \ge b) \le P(\sum_{i=1}^{n}\eta_i - \underline{\eta} \ge b) \le P(\sum_{i=1}^{n}\eta_i \ge b),
\label{eq:relationship}
\end{equation}
% \vskip -0.1in
\normalsize
which indicates $p_{\eta r} \le p_{\underline{\eta}} \le p_{o}$. For the filters in the $\eta$ layer, we naturally assume that the contribution of a filter to the network's performance cannot be infinite, i.e., the variances of filters' contributions are uniformly bounded.
\begin{equation}
\exists C_1 >0,~~~\text{s.t.}~~~ \mathbb{D}\eta_i \le C_1, i=1,2,\cdots,n.
\label{eq:a1}
\end{equation}

By Chebyshev's inequality, for any real number $\epsilon>0$,
\small
% \vskip -0.1in
\begin{equation}
P(\frac{1}{n}|\sum_{i=1}^n(\eta_i-\mathbb{E}\eta_i)|\ge \epsilon) \le \frac{\mathbb{D}(\sum_{i=1}^n\eta_i)}{\epsilon^2n^2}.
\label{eq:cheb}
\end{equation}
\normalsize
% \vskip -0.1in
With Equation \eqref{eq:a1}, it is obvious that we have $\Cov(\eta_i,\eta_j)\le\sqrt{\mathbb{D}\eta_i \cdot \mathbb{D}\eta_j}\le C_1$.

We further define that there are $C_2n~(0 \le C_2 \le 1)$ pairs of correlated filters in the $\eta$ layer, i.e.,
$\#\{(i,j): \Cov (\eta_i, \eta_j) \neq 0, i\neq j,\ i,j=1,\cdots,n.\} \le C_2n$. Then we have,
\small
% \vskip -0.1in
\begin{equation*}
\begin{aligned}
\mathbb{D}(\sum_{i=1}^n\eta_i) &=\sum_{i=1}^n \mathbb{D}\eta_i + \sum_{i \neq j}\Cov(\eta_i,\eta_j) \\
&\le C_1n +C_1C_2n = C_1(1+C_2)n.
\end{aligned}
\end{equation*}
\normalsize
By Equation \eqref{eq:cheb},
% \vskip -0.1in
\small
\begin{equation*}
P(\frac{1}{n}|\sum_{i=1}^n(\eta_i-\mathbb{E}\eta_i)|\ge \epsilon) \le \frac{C_1(1+C_2)}{\epsilon^2n} \to 0.
\end{equation*}

This means $\frac{1}{n} \sum_{i=1}^n(\eta_i-\mathbb{E}\eta_i)$ converges in probability to zero, i.e., $\frac{1}{n}\sum_{i=1}^n(\eta_i - \mathbb{E}\eta_i) \xrightarrow[]{P} 0$. Suppose the number of filters in the $\eta$ layer $n$ is large enough, say $n>\frac{2b}{\epsilon_0}$. 

We consider that a filter's contribution needs to be positive, but it could be infinitely small, i.e., the expectation of filters' contributions have a uniform positive lower bound.
\begin{equation}
\exists \epsilon_0 > 0,~~~\text{s.t.}~~~\mathbb{E}\eta_i \ge \epsilon_0, i=1,2,\cdots,n.
\label{eq:a3}
\end{equation}
With Equation \eqref{eq:a3}, we have,
\small
% \vskip -0.25in
\begin{equation*}
\begin{aligned}
&~~~P(\frac{1}{n}\sum_{i=1}^n(\eta_i-\mathbb{E}\eta_i)>-\frac{\epsilon_0}{2}) = P(\sum_{i=1}^n\eta_i > \sum_{i=1}^n\mathbb{E}\eta_i - \frac{\epsilon_0}{2}n) \\
&=P \Big{(} \sum_{i=1}^n\eta_i > \frac{\epsilon_0}{2}n + \sum_{i=1}^n (\mathbb{E}\eta_i - \epsilon_0) \Big{)} \\
&\le P(\sum_{i=1}^n\eta_i > \frac{\epsilon_0}{2}n) \le P(\sum_{i=1}^n\eta_i > b).
\end{aligned}
\end{equation*}
% \vskip -0.1in
\normalsize
Letting $n\to+\infty$, taking the limit and note that   $\frac{1}{n}\sum_{i=1}^n(\eta_i - \mathbb{E}\eta_i) \xrightarrow[]{P} 0$, we have
\small
% \vskip -0.1in
\begin{equation*}
\lim_{n\to\infty}P(\sum_{i=1}^n\eta_i > b) \ge \lim_{n\to\infty} P(\frac{1}{n}\sum_{i=1}^n(\eta_i-\mathbb{E}\eta_i)>-\frac{\epsilon_0}{2}) = 1,
\end{equation*}
% \vskip -0.15in
\normalsize
% Similarly,
% \vskip -0.1in
\small
\begin{equation*}
\lim_{n\to\infty}P(\sum_{i=1}^n\eta_i - \eta_r > b) = \lim_{n\to\infty}P(\sum_{i=1}^n\eta_i - \underline{\eta} > b) = 1,
\end{equation*}
\normalsize
and then we have $p_{\eta r} \approx p_{\underline{\eta}} \approx p_{o}$ for $n$ large enough. Note that $p_{\underline{\xi}} \le p_o \approx p_{\underline{\eta}}$ and observe that $p_g$ is the weighted average of $p_{\underline{\xi}}$ and $p_{\underline{\eta}}$. Hence we have $p_{\underline{\xi}} \le p_g \le p_{\underline{\eta}}$. It is worth mentioning that we cannot imply $p_g \approx p_{\eta}$ from Equation \eqref{eq:gl} by letting $n \rightarrow \infty$ because we do not assume $m/n \rightarrow 0$.

In summary, we have $p_{\underline{\xi}} \le p_{g} \le p_{\eta r} \le p_{\underline{\eta}} \le p_{o}$, which indicates that (even randomly) pruning a filter in the layer with much larger redundancy outperforms pruning the least important filter across all layers. Here we consider the CNN as a black-box and we do not assume any prior distribution for the RVs to achieve a good generalization. So the conclusion holds no matter how the RVs are distributed. Indeed, the conclusion relies on the assumption $n\to+\infty$. However, the assumption can be relaxed in real world applications such that $p_{\underline{\eta}} \ge p_{g}$ still holds on average (though not in every filter selection step). Appendix A presents some intuitive examples on a number of networks, which empirically provides evidence to validate the analysis above. As shown in Appendix A, the number of filters in a redundant layer does not need to be very large. Even when we use this naive strategy (randomly pruning filters in the layer with the most number of filters) with a standard AlexNet, it outperforms a number of popular pruning approaches. As we can see in the following, for more sophisticated architectures that contain less redundancy, such as ResNet, with a well designed metric to measure the layer redundancy, our proposed approach that prunes the least important filters in the layer(s) with larger redundancy outperforms recent pruning approaches that removes the least important filters across all layers.

% In practice, we know that $n\to+\infty$ does not hold and $p_{\eta r} \ge p_{g}$ is not always true. However, if the $\eta$ layer is with larger redundancy than other layers, it is still more likely that $p_{\underline{\eta}} \ge p_{g}$, which indicates we can prune the least important filters in the layer(s) with larger redundancy to achieve better performance than pruning least important filters across all layers.

%\footnote{When the redundancy of a layer is largely reduced during the pruning procedure, $v_{gl} \le v_{sr}$ does not hold any more. For example, suppose that $m=n$ and $a=b$, we have $P(\sum_{i=1}^n\eta_i \ge a)=P(\sum_{i=1}^m\xi \ge b)$, and $P(\sum_{i=1}^n\eta_i - \underline{\eta} \ge a)=P(\sum_{i=1}^m\xi - \underline{\xi} \ge b)$. Therefore, $v_{gl}=P(\sum_{i=1}^m\xi \ge b)+P(\sum_{i=1}^n\eta_i - \underline{\eta} \ge a)=v_{sl}=v_{fl} \ge v_{sr}$.}.

% \vskip -0.1in
\begin{figure*}[htb]
% \vskip -0.1in
\centering
    \includegraphics[width=0.98\textwidth]{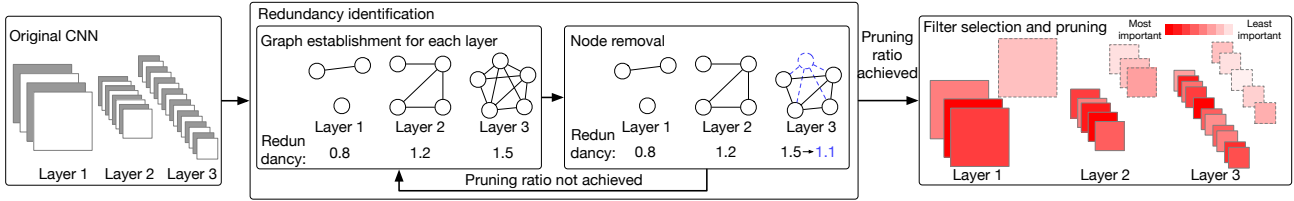}
% \vskip -0.1in
\caption{Overall workflow of the proposed approach. The number below each graph refers to the measurement of redundancy, which are just used for the illustration purpose and do not reflect the real measurements of the graphs.}
\label{fig:whole_pic}
\vskip -0.1in
\end{figure*}

\section{Methodology}
\label{sec:method}
% We first introduce the notations, preliminaries of network pruning and graph-related concepts used in our approach. Then we describe the proposed approach based on structural redundancy reduction.

\subsection{Notations and preliminaries}
\label{subsec:notation}
\noindent{\bf Network pruning.}
Suppose a CNN has $L$ layers. For the $i$-th layer, the number of the input and output channels are represented by $N_i$ and $N_{i+1}$. Therefore, the CNN's parameters $\bf W$ can be represented as $\{{\bf W}^{(i)} \in \mathbb{R}^{N_{i+1} \times N_i \times h_i \times w_i},~i=1,2,\cdots,L\}$, where $h_i$ and $w_i$ are the filter's width and height. Channel pruning is formulated to find a set of parameters $\bf W'$ optimized on certain objective functions such that $||{\bf W'}||_0 < K$, where $||\cdot||_0$ denotes the $\ell_0$ norm and $K$ limits the number of non-zero filters in $\bf W'$. Depending on different configurations, the objective functions can be minimizing a CNN's cost function, drop of training accuracy, or the reconstruction error, etc.
% Let ${\bf w}^{(i)}_j \in \mathbb{R}^{N_i \times h_i \times w_i}$ denote the $j$-th filter in the $i$-th layer, where $h_i$ and $w_i$ are the filter's width and height.

\noindent{\bf Graph theory.} 
Let $X$ be a finite set. An undirected graph is a pair $(X,E)$, where $E$ is a symmetric subset of $X \times X \setminus \{(x,x): \ x\in X\}$. We call $x\in X$ a vertex (or a node) and $(x,y)\in E$ an edge. For $x,y \in X$, a path from $x$ to $y$ is a finite sequence $\{x_0, x_1, \cdots, x_n\} \subset X$ such that $x_0 = x, \ x_n = y$ and $(x_i, x_{i+1}) \in E$. In general, the above path may not be unique if such a path exists. Denote $d(x,y)$ the minimal length of paths from $x$ to $y$ if $x$ and $y$ can be connected by a path; $d(x,y)=0$ if $x=y$; and $d(x,y) =+\infty$ if $x$ and $y$ cannot be connected by a path. Then it is clear that $d(x,y)$ is an integer value metric. Recall that the degree of a vertex $x\in X$ is the total number of edges connected to $x$, i.e., $\deg(x) = \# \{(x,y): \ (x,y) \in E\}$ ($\#A$ is the total number of elements in $A$).

\subsection{Pruning with structural redundancy reduction}
\label{subsec:detailmethod}
{\bf Overall architecture.} We showed that pruning filters in the layer with larger redundancy outperforms pruning the least important filters across all layers. Our approach focuses on measuring how much redundancy exists in each layer and pruning filters from the most redundant layer(s) (Fig.~\ref{fig:whole_pic}). To measure the structural redundancy in a network, for each layer, we build an undirected graph in which each vertex represents a filter and the edges are defined with the distances between filter weights. We use two quantities associated with the graph, i.e., quotient space size and $\ell$-covering number, as a measurement of how much redundancy exists in each graph, which is considered as the redundancy exists in each layer. At each time step, after the graph establishment and redundancy quantification, we randomly remove a vertex and its associated edges from the graph identified as with the most redundancy. Then we recalculate the redundancy after graph reconstruction for the next iteration. This process continues until a target is reached (e.g., a certain number of filters are pruned). Finally, we prune the filters in each layer according to the remaining number of vertices in each graph with a certain filter selection criterion. Note that in the filter pruning phase, filters are ranked separately in each layer, rather than globally across all layers. Since the redundancy identification phase has selected a different number of filters in each layer, our approach is a layer-adaptive approach.

We present the details of our approach as follows: graph establishment, calculation of quotient space size and $\ell$-covering number, intuition and quantification of graph redundancy, and filter selection.

{\bf Graph establishment. } To illustrate how to build a graph for a convolutional layer, we use $X$ to represent the filter weights of a certain layer ${\bf W}^{(i)}$ for simplicity. We first flatten and normalize the filter weights, which changes their lengths to $1$. After that $X$ becomes a finite subset of $n$ dimensional unit sphere ${\mathbb S}^n = \{x \in {\mathbb R}^n: |x|=1\}$ in ${\mathbb R}^n$, where $n=N_i \times h_i \times w_i$ and $|x|$ is the length of $x$ in ${\mathbb R}^n$. We define a graph on $X$ as follows (assuming the elements in $X$ are distinct). We choose a positive real number $\gamma >0$ and define an edge set on $X$ as 
% \vskip -0.1in
\[
E = \{(x,y) \in X \times X \setminus \Delta: |x-y| / \sqrt{n} \le \gamma \},
\]
% \vskip -0.1in
where $\Delta = \{(x,x): x \in X\}$ is the diagonal of $X$, and $|x - y|$ is the Euclidean distance on ${\mathbb R}^n$.
Then we get a graph $(X,E)$. By definition $(x,y) \in E$ implies $x$ and $y$ are approximately equal if $\gamma$ is small. 

{\bf $\ell$-covering number.} Recall that $(X,d)$ is a metric space, where $d$ is the graph metric defined previously. Let $\ell >0$ be a fixed natural number, a subset $X_0 \subset X$ is called an $\ell$-cover set of $X$, if  $X \subset \bigcup\{ B(x', \ell): x'\in X_0\}$, where $B(x',\ell) = \{x\in X: \ d(x',x) \le \ell\}$ is the ball centered at $x'$ with radius $\ell$. This means $X$ is covered by the balls $\{B(x',\ell): \ x'\in X_0\}$. We call the following quantity the $\ell$-covering number of $X$:
% \vskip -0.2in
\[
N_\ell^c=(N_\ell^c(X) =) \min\{\# X_0:X_0 \mbox{ is an } \ell \mbox{-cover set of } X\}.
\]
% \vskip -0.1in
%where $\# X_0$ is the total number of elements in $X_0$.

{\bf Decomposition of a graph.} We call a graph connected if for any $x \not= y$, there exists a path from $x$ to $y$. In this case $d(x,y) < \infty$ for all $x,y \in X$. For an unconnected graph $(X,E)$, we define the notation ``$\sim$" on $X$ as follows: $x\sim y$ if and only if there exists a path from $x$ to $y$. Then it is clear that ``$\sim$" is an equivalence relation. Let $X/ \sim =\{X_1, X_2, \cdots, X_k\}$
% \vskip -0.2in
% \begin{equation} \label{eq1}
% X/ \sim =\{X_1, X_2, \cdots, X_k\}
% \end{equation}
% \vskip -0.1in
be the quotient space. This mathematical concept means that: using an equivalence relation, we can decompose the set $X$ as a disjoint union $X = X_1 \cup X_2 \cup \cdots \cup X_k$ such that the elements in the same $X_i$ are equivalent. We call the number $k$ (the total number of equivalence classes) the quotient space size. Intuitively, $k$ is the number of unconnected sub-graphs of $(X,E)$. 
% We give a straightforward example of the calculation of the above two quantities (Fig.~\ref{fig:graph_eg}). First of all, there are $3$ unconnected sub-graphs in total, so $k=3$. For each sub-graph, $1$, $1$, and $2$ balls (centered at the darker nodes) need to be selected as the $1$-cover set to cover all the vertices in the graph, so $N_1^c=4$.
% \vskip -0.15in
% \begin{figure}[htb]
% \centering
%     \includegraphics[width=0.3\textwidth]{figs/graph_illustration2_nips.eps}
% \vskip -0.05in
% \caption{An example of the calculation of $\ell$-covering number and quotient space size. Darker color nodes indicate the selected ball centers and lighter color nodes represent the elements within the corresponding balls.}
% \label{fig:graph_eg}
% \vskip -0.1in
% \end{figure}

\noindent{\bf Graph redundancy, intuition and quantification.} Intuitively, larger values of the quotient space size and $\ell$-covering number indicate a more complicated set of data (with less redundancy). In fact, $x\in B(x',\ell)$ if and only if $d(x,x') \le \ell$, so $x$ and $x'$ are approximate equal. Hence the covering number can be approximately considered as the total number of vectors in $X$ that are linearly independent. In our implementation we simply use $\ell =1$, with the consideration of both performance and computation efficiency. Based on the above analysis, we define the graph (layer) redundancy as in Equation~\eqref{eq:redun}.
% \vskip -0.2in
\begin{equation}
R(X) = \frac{N}{w_1 k + w_2 N_1^c},
\label{eq:redun}
\end{equation}
% \vskip -0.1in
where $\{w_1, w_2\}$ is a probability weight that balances the importance of $k$ and $N_1^c$, $N$ is the number of filters. Besides the graph redundancy, we also investigate other criteria (i.e., the number of filters and principal component analysis (PCA)) to measure the structural redundancy in the ablation study.
% {\color{red}It can be guaranteed that if there's only 1 filter in a layer, it will not be selected, unless all layers contain only 1 filter. derivations ...}

\noindent{\bf Estimate of the 1-covering number.} Since the calculation of $\ell$-covering number is NP-hard and time-consuming in practice \cite{gibson2010clusters}, we propose a lightweight method to estimate $N_1^c$. Let $X_0$ be the $1$-cover set of a graph $X$, such that $\# X_0 = N_1^c$. 
%\vskip -0.2in
%\[
%{\color{orange}(N_1^c=\#\mathscr{B}_0?)}\#\mathscr{B}_0 = \min\{\#\mathscr{B}: \mathscr{B} \text{ %is an $1$-cover set of }G\},
%\]
% \vskip -0.1in
We estimate $\# X_0$ as follows. Fix an integer $\ell(=1 \text{ or } 2)$ and let $x_1^{(\ell)} \in X$, s.t. $\deg(x_1^{(\ell)}) = \max\{\deg(x):~x \in X\}$.
% \vskip -0.2in
% \[
% \deg(x_1^{(\ell)}) = \max\{\deg(x):~x \in X\}.
% \]
% \vskip -0.1in
We define a finite sequence $\{x_1^{(\ell)},x_2^{(\ell)},\cdots,x_{n_{\ell}}^{(\ell)}\}$ by induction: If we have defined $x_k^{(\ell)}$, then there are two possible cases: (i) $X =\bigcup_{i=1}^{k} B(x_i^{(\ell)}, \ell)$, i.e., the family of balls $\{B(x_i^{(\ell)}, \ell):~1 \le i \le k\}$ is an $\ell$-cover of $X$. Then we stop the construction of the sequence and get $\{x_1^{(\ell)},x_2^{(\ell)},\cdots,x_{n_k}^{(\ell)}\}$. (ii) Otherwise, choose (any) $x_{k+1}^{(\ell)} \in X \backslash \bigcup_{i=1}^{k}B(x_i^{(\ell)}, \ell)$, s.t.
% \vskip -0.25in
\[
\deg(x_{k+1}^{(\ell)})=\max\{\deg(x):~x \in X \backslash \bigcup_{i=1}^{k}B(x_i^{(\ell)}, \ell)\}.
\]
% \vskip -0.1in
We repeat the above process eventually, and get the sequence
% \vskip -0.1in
\[
\{x_1^{(\ell)},x_2^{(\ell)},\cdots,x_{n_{\ell}}^{(\ell)}\},~~~~l=1\text{ or }2.
\]
% \vskip -0.05in
% It is clear that:
% \begin{itemize}
% \itemsep0em
% \item $\#\mathscr{B}_0 \le n_1$. Since the family $\{B(x_k^{(1)},1):~1 \le k \le n_1\}$ is a $1$-cover of $G$.
% \item $\#\mathscr{B}_0 \ge n_2$. Indeed, for any $i \ne j~(i,j \le n_2)$, we have $d(x_i^{(2)},x_j^{(2)}) \ge 3$, where $d(\cdot,\cdot)$ is the distance on the graph. For each $B \in \mathscr{B}_0$, and any $x,y \in B$, we have $d(x,y) \le 2$. $\mathscr{B}_0$ is a cover of $G$. Hence, for any $x_i^{(2)}$, there exists (may not unique) $B^i \in \mathscr{B}_0$, s.t. $x_i^{(2)} \in B^i$. If $i \ne j$, then $B^i \ne B^j$ (otherwise, $d(x_i^{(2)},x_j^{(2)}) \le 2$). We see that $n_2 \le \#\mathscr{B}_0$.
% \end{itemize}
It is obvious that we have $N_1^c = \# X_0 \le n_1$ because the family $\{B(x_k^{(1)},1):~1 \le k \le n_1\}$ is a $1$-cover of $X$. Moreover, for any $i \ne j~(i,j \le n_2)$, we have $d(x_i^{(2)},x_j^{(2)}) \ge 3$. On the other hand, for each $x_0 \in X_0$, and any $x,y \in B(x_0,1)$, we have $d(x,y) \le d(x,x_0)+ d(x_0,y) \le 2$. Recall that $X_0$ is a 1-cover set of $X$, then for any $x_i^{(2)}$, there exists (may not unique) $x_0 \in X_0$ such that $x_i^{(2)} \in B(x_0,1)$. Moreover, for $i \ne j$, $x_i^{(2)}$ and $x_j^{(2)}$ cannot be in the same ball $B(x,1)$ (otherwise $d(x_i^{(2)}, x_j^{(2)}) \le 2$, a contradiction). We see $n_2 \le \# X_0 = N_1^c$. Hence $n_2 \le N_1^c \le n_1$.

\begin{table*}[t]
% \vskip -0.1in
\centering
\begin{tabular}{|c|c|c|c|c|c|}
\hline
Model & Approach & Acc. before prune & Acc. after prune & Acc. drop & FLOPs drop \\
\hline
\multirow{5}{*}{ResNet20} & MW & 92.35\% & 90.93\% & 1.42\% & 41.0\%\\
& SFP & 92.20\% & 90.83\% & 1.37\% & 42.4\% \\
% \cline{2-6} 
& GM & 92.20\% & 91.09\% & 1.11\% & 42.2\% \\
& TAS  & - & 92.88\% & 0.00\% & 45.0\% \\
% \cline{2-6} 
% \hhline{*{1}{|~}*{5}{|-}|}
& \cellcolor{Gray}SRR-GR & \cellcolor{Gray}92.27\% & \cellcolor{Gray}92.48\% & \cellcolor{Gray}\bf -0.21\% & \cellcolor{Gray}\bf 45.8\% \\
% & Graph Redundancy (Ours) & 92.35\% & 91.35\% & \bf 1.00\% & \bf 45.8\% \\
% \hline
\Xhline{1.8\arrayrulewidth}
\multirow{9}{*}{ResNet56} & MW & 93.51\% & 92.90\% & 0.61\% & 51.5\% \\
& NISP & - & 93.01\% & - & 35.5\% \\
& GAL & 93.26\% & 93.38\% & -0.12\% & 37.6\% \\
& DCP & 93.80\% & 93.49\% & 0.31\% & 49.8\% \\
& HRank & 93.26\% & 93.17\% & 0.09\% & 50.0\% \\
& SCP & 93.69\% & 93.23\% & 0.46\% & 51.5\% \\
& SFP & 93.59\% & 92.26\% & 1.33\% & 52.6\% \\
% \cline{2-6} 
& GM & 93.59\% & 92.93\% & 0.66\% & 52.6\% \\
% \cline{2-6} 
& TAS & - & 93.69\% & 0.77\% & 52.7\% \\
% \cline{2-6} 
% \hhline{*{1}{|~}*{5}{|-}|}
& \cellcolor{Gray}SRR-GR & \cellcolor{Gray}93.38\% & \cellcolor{Gray}93.75\% & \cellcolor{Gray}\bf -0.37\% & \cellcolor{Gray}\bf 53.8\% \\
% & Graph Redundancy (Ours) & 93.51\% & 93.04\% & \bf 0.47\% & \bf 53.8\% \\
\hline
\end{tabular}
% \vskip -0.1in
\caption{Results of ResNet on CIFAR-10. MW results is with our own implementation. GR is graph redundancy.}
\label{tab:cifar_single}
\vskip -0.2in
\end{table*}
We can use $\tilde{N_1^c}=\frac{1}{2}(n_1+n_2)$ to estimate $N_1^c$, if $|n_1-n_2|$ is acceptably small. Although we cannot theoretically find its upper bound, extensive experiments on various networks show that $\tilde{N_1^c}$ is good enough as an estimation of $N_1^c$, and the computing time of $\tilde{N_1^c}$ is negligibly small (see the Analysis and ablation study section).

\noindent{\bf Filter selection strategy.} After identifying the layers with large redundancy, we prune unimportant filters from these layers. We can either train a pruned network architecture from scratch with random initialization or prune certain filters from the pre-trained network and do fine-tuning. There are various approaches for unimportant filter selection. In our study, we use a very common and simple strategy, i.e., pruning the filters with smaller absolute weights \cite{li2016pruning}. This method avoids feeding a large number of training samples into the CNN to get filter rankings, which is usually computationally intensive \cite{polyak2015channel, molchanov2016pruning}. But in general, our approach can be used together with any filter selection criterion.

\section{Experiments}
\label{sec:exp}
\subsection{Experiment settings}
% We evaluate our proposed approach with the single-shot pruning scheme (pruning a large number of filters at one time), with ResNet \cite{he2016deep} on two widely-used image classification datasets, i.e., CIFAR-10 \cite{krizhevsky2009learning} and ImageNet ILSVRC-2012 \cite{deng2009imagenet}.
We first evaluate our approach with the single-shot pruning scheme (pruning a large number of filters at one time), with two-widely used benchmark datasets (CIFAR-10 \cite{krizhevsky2009learning} and ImageNet ILSVRC-2012 \cite{deng2009imagenet}) on ResNet. We also present the results with the progressive pruning scheme (pruning a small number of filters and fine-tuning the remaining network for mutiple times), which are presented in the Appendix due to the space limitation.

% architectures (AlexNet \cite{krizhevsky2012imagenet}, VGG16 \cite{simonyan2014very}, and ResNet \cite{he2016deep}), with three widely-used benchmark datasets (CIFAR-10 \cite{krizhevsky2009learning}, Birds-200 \cite{birds200}, and ImageNet ILSVRC-2012 \cite{deng2009imagenet}). We evaluate the performance on two pruning schemes, namely, single-shot pruning (pruning a large number of filters at one time), and progressive pruning (pruning a small number of filters and fine-tuning the remaining network, for multiple iterations). However, due to space limitations, results on progressive pruning are presented in Appendix. The number of filters, parameters, and floating point operations (FLOPs) are used to quantify network compression. 

% We compare the performance of our approach with several recent channel pruning methods, namely, minimum weight (MW) \cite{li2016pruning}, mean activation \cite{polyak2015channel}, Taylor expansion \cite{molchanov2016pruning}, ApoZ \cite{hu2016network}, more is less (MIL) \cite{dong2017more}, soft filter pruning (SFP) \cite{he2018soft}, geometric median (GM) \cite{he2019filter}, and principal filter analysis (PFA) \cite{suau2018network}. We also plot the performance of randomly pruning filters as a reference in progressive pruning experiments.
For single-shot pruning, we use ResNet\{20,56\} on the CIFAR-10 dataset and Resnet50 on ImageNet to evaluate the performance, in terms of accuracy drop and FLOPs reduction. We used the widely-used ResNet architecture as described in \cite{he2016deep}. For the CIFAR-10 experiments, the models are trained following the setup in \cite{he2019filter}. For the ImageNet experiments, pre-trained models from torchvision are used. We first evaluate the layer redundancy in the pre-trained models and identify the number of filters to be pruned in each layer, with our proposed approach. Then we prune the filter in each layer with the corresponding numbers identified and fine-tune the slimmed network. We follow the fine-tuning strategy in \cite{he2018soft}. For CIFAR-10, we fine-tune each  pruned network for 200 epochs, with a learning rate starting from 0.1, which is divided by 10 at the epochs 60, 120, and 160. For ImageNet, we fine-tune each pruned network for 150 epochs, with a learning rate starting from 0.1, which is divided by 10 every 30 epochs. 
For all the models, we use an SGD optimizer with a momentum of 0.9, a weight decay of $2e^{-5}$, and a batch size of 256. For the graph associated parameters, we use $w_1=0.35$, $w_2=0.65$ to emphasize the importance of $\ell$-covering number. We choose $\gamma=0.034$ to achieve the best performance. We implement the experiments with Pytorch 1.3 \cite{paszke2019pytorch}.

We compare the performance of our approach with several recent channel pruning methods, namely, minimum weight (MW) \cite{li2016pruning}, Taylor expansion \cite{molchanov2016pruning}, average percentage of zero activation neurons (APoZ) \cite{hu2016network}, soft filter pruning (SFP) \cite{he2018soft}, discrimination-aware channel pruning (DCP) \cite{zhuang2018discrimination}, neuron importance score propagation (NISP) \cite{yu2018nisp}, slimmable neural networks (SNN) \cite{yu2018slimmable}, autopruner (AP) \cite{luo2018autopruner}, generative adversarial learning (GAL) \cite{lin2019towards}, geometric median (GM) \cite{he2019filter}, transformable architecture search (TAS) \cite{dong2019network}, cluster pruning (CUP) \cite{duggal2019cup}, ABC \cite{lin2020channel}, trained rank pruning (TRP) \cite{xu2020trp}, soft channel pruning (SCP) \cite{kang2020operation}, and high-hank (HRank) \cite{lin2020hrank}.

\subsection{Performance evaluation}
% We use ResNet-\{20,56\} on the CIFAR-10 dataset and Resnet-\{18,34,50\} on ImageNet to evaluate the performance with single-shot pruning, in terms of accuracy drop and FLOPs reduction. We fine-tune each pruned network for 120 epochs, with a learning rate starting from 0.1, which is decayed a factor 10 every 30 epochs. 

\begin{table*}[t]
% \vskip -0.12in
\centering
\begin{tabular}{|C{1.555cm}|C{1.5cm}|C{1.5cm}|C{1.6cm}|C{1.6cm}|C{0.9cm}|C{0.9cm}|C{0.88cm}|}
\hline
\multirow{2}{*}{Approach} & Top-1 acc.  & Top-5 acc. & Top-1 acc. & Top-5 acc. & Top-1 & Top-5 & FLOPs  \\
& baseline & baseline & after prune & after prune & acc. $\downarrow$ & acc. $\downarrow$ & $\downarrow$ \\
% \hline
\Xhline{1.8\arrayrulewidth}
MW & 76.13\% & 92.86\% & 71.24\% & 90.38\% & 4.89\% & 2.48\% & 41.8\%\\
% \cline{1-9} 
% ApoZ \cite{hu2016network} & 76.13\% & 92.86\% & 74.97\% & 92.30\% & 1.16\% & 0.56\% & 41.7\% \\
% \cline{1-9} 
% Taylor \cite{molchanov2016pruning} & 76.13\% & 92.86\% & 74.78\% & 92.38\% & 1.35\% & 0.48\% & 40.7\% \\
% \cline{1-9} 
SFP &  76.15\% &  92.87\% & 74.61\% & 92.06\% & 1.54\% & 0.81\% & 41.8\% \\
% \cline{1-9} 
GM &  76.15\% &  92.87\% & 75.03\% & 92.40\% & 1.12\% & 0.47\% & 42.2\% \\
% SSS  &  - & - & 71.82\% & 90.79\% & - & - & 43.0\% \\
GAL&  76.15\% & 92.87\% & 71.95\% & 90.94\% & 4.20\% & 1.93\% & 43.0\% \\
TAS &  - & - & 76.20\% & 93.07\% & 1.26\% & 0.48\% & 43.5\% \\
HRank & 76.15\% & 92.87\% & 74.98\% & 92.33\% & 1.17\% & 0.54\% & 43.8\% \\
SNN  &  - &  - & 74.90\% & - & 1.10\% & - & 43.9\% \\
% \cline{2-9}
% \cline{1-9} 
% \hhline{*{1}{|~}*{8}{|-}|}
\cellcolor{Gray}SRR-GR & \cellcolor{Gray}76.13\% & \cellcolor{Gray}92.86\% &  \cellcolor{Gray}75.76\% &  \cellcolor{Gray}92.67\% & \cellcolor{Gray}\bf 0.37\% & \cellcolor{Gray}\bf 0.19\% & \cellcolor{Gray}\bf 44.1\%\\
TRP & - & - & 74.06\% & 92.07\% & - & - & 44.4\% \\
AP  &  76.15\% &  92.87\% & 74.76\% & 92.15\% & 1.39\% & 0.72\% & 51.2\% \\
GM &  76.15\% &  92.87\% & 74.13\% & 91.94\% & 2.02\% & 0.93\% & 53.5\% \\
ABC  & 76.01\% & 92.96\% & 73.86\% & 91.69\% & 2.15\% & 1.27\% & 54.0\% \\
SCP & 75.89\% & 92.98\% & 74.20\% & 92.00\% & 1.69\% & 0.98\% & 54.3\% \\
CUP  &  - &  - & - & - & 1.47\% & 0.88\% & 54.5\% \\
GAL &  76.15\% & 92.87\% & 71.80\% & 90.82\% & 4.35\% & 2.05\% & 55.0\% \\
\cellcolor{Gray}SRR-GR & \cellcolor{Gray}76.13\% & \cellcolor{Gray}92.86\% &  \cellcolor{Gray}75.11\% &  \cellcolor{Gray}92.35\% & \cellcolor{Gray}\bf 1.02\% & \cellcolor{Gray}\bf 0.51\% & \cellcolor{Gray}\bf 55.1\%\\
\hline
\end{tabular}
\caption{ResNet50 results on ImageNet. Results of MW, APoZ, and Taylor are based on our own implementation.}
\label{tab:imagenet_single}
\vskip -0.1in
\end{table*}

\noindent{\bf CIFAR-10.} Results of pruning ResNet20 and ResNet56 on CIFAR-10 are presented in Table~\ref{tab:cifar_single}. Our approach prunes a large percent of FLOPs from both architectures without performance degradation, which outperforms the previous state-of-the-art with an obvious margin. For ResNet20, we prune $45.8\%$ FLOPs and the test accuracy is increased by $0.21\%$. Our pruned ResNet56 model reduces $53.8\%$ FLOPs and achieves a test accuracy of $93.75\%$, which outperforms the baseline by $0.37\%$.

\noindent{\bf ImageNet.} Results of ResNet50 on ImageNet are shown in Table~\ref{tab:imagenet_single}. Our pruned model with $44.1\%$ FLOPs reduction only loses $0.37\%$ top-1 accuracy and $0.19\%$ top-5 accuracy. When pruning comparable FLOPs, the top-1 accuracy of the previous state-of-the-art approaches usually drop by more than $1\%$. As the pruning ratio increases to $55.1\%$, the proposed approach can still achieve a promising test accuracy ($1.02\%$ and $0.51\%$ drop for top-1 and top-5 accuracy), which is the best performance compared with recent works. These results verify the effectiveness of our approach on the single-shot pruning scheme. It is worth mentioning that the MW approach can be considered as a baseline of our approach because we add a redundancy identification stage before using MW for filter pruning. It is observed that pruning filters uniformly with MW results in unsatisfactory results ($71.24\%$ top-1 accuracy). After identifying the redundancy and pruning the corresponding number of filters in each layer, the performance is significantly improved by around $4\%$.

\begin{table}[htb]
\centering
\begin{tabular}{|c|c|c|c|}
\hline
{Num of} & {1-covering} & {Time with} & {Time with} \\
filters & number & oracle method & our method \\
\hline
\multirow{4}{*}{64} & 1 & 0.001 & 0.0015\\
\cline{2-4} 
& 2 & 0.045 & 0.0013\\
\cline{2-4} 
& 3 & 1.384 & 0.0020\\
\cline{2-4} 
& 4 & 28.21 & 0.0026\\
\hline
\multirow{3}{*}{192} & 1 & 0.023 & 0.0014\\
\cline{2-4} 
& 2 & 0.974 & 0.0027\\
\cline{2-4} 
& 3 & 90.29 & 0.0028\\
\hline
\end{tabular}
% \vskip -0.1in
\caption{Average computation time of the oracle and our proposed method for the 1-covering number calculation/estimate (in secs).}
\label{tab:time}
\vskip -0.1in
\end{table}
\section{Analysis and ablation study}
\label{sec:ana}

\subsection{l-covering number estimate and computation time}
\begin{figure*}[t]
\centering
\begin{minipage}{2\columnwidth}
\centering
\subfigure[AlexNet]{
\includegraphics[width=0.23\textwidth]{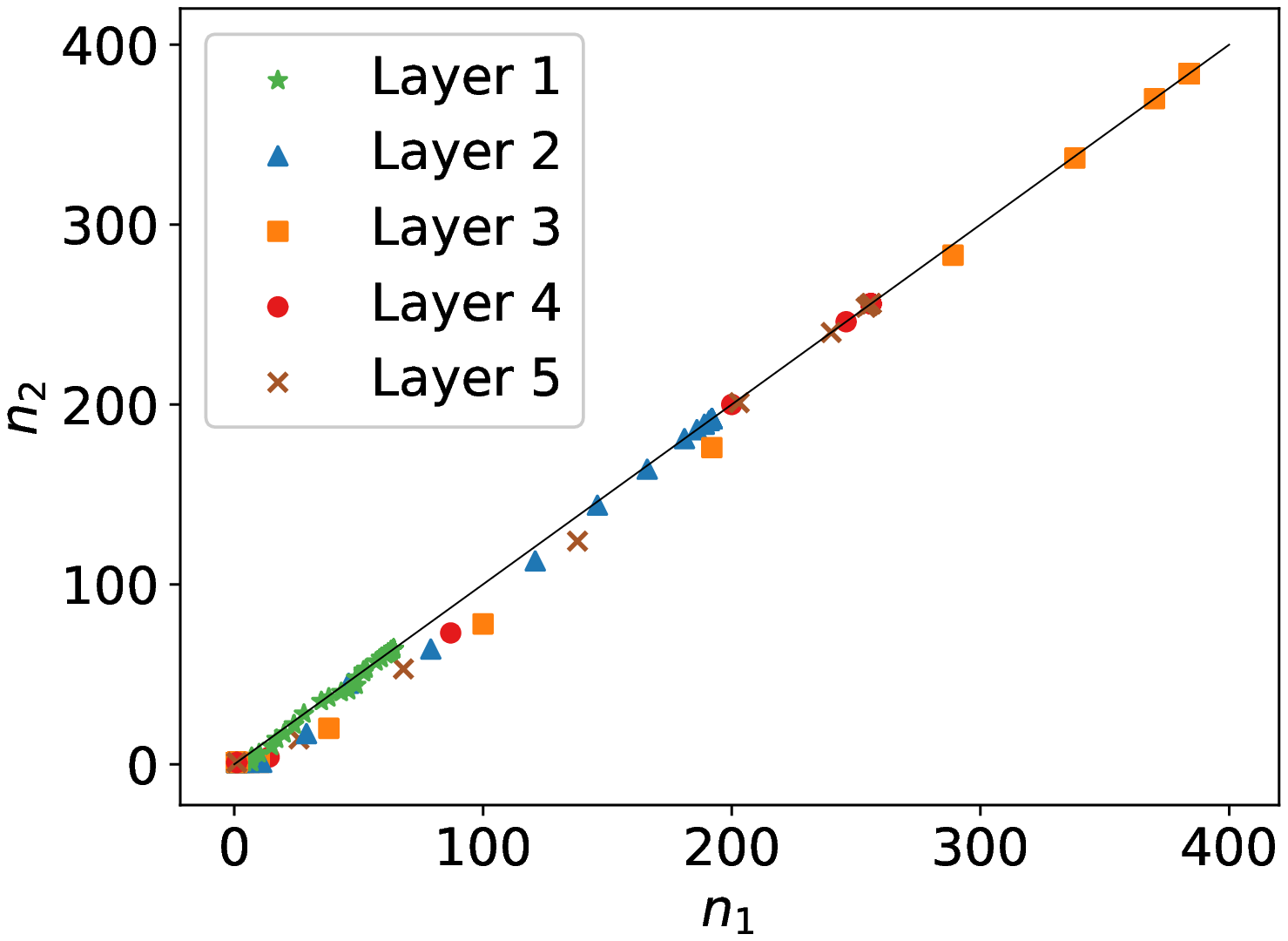}}
\subfigure[VGG16]{
\includegraphics[width=0.23\textwidth]{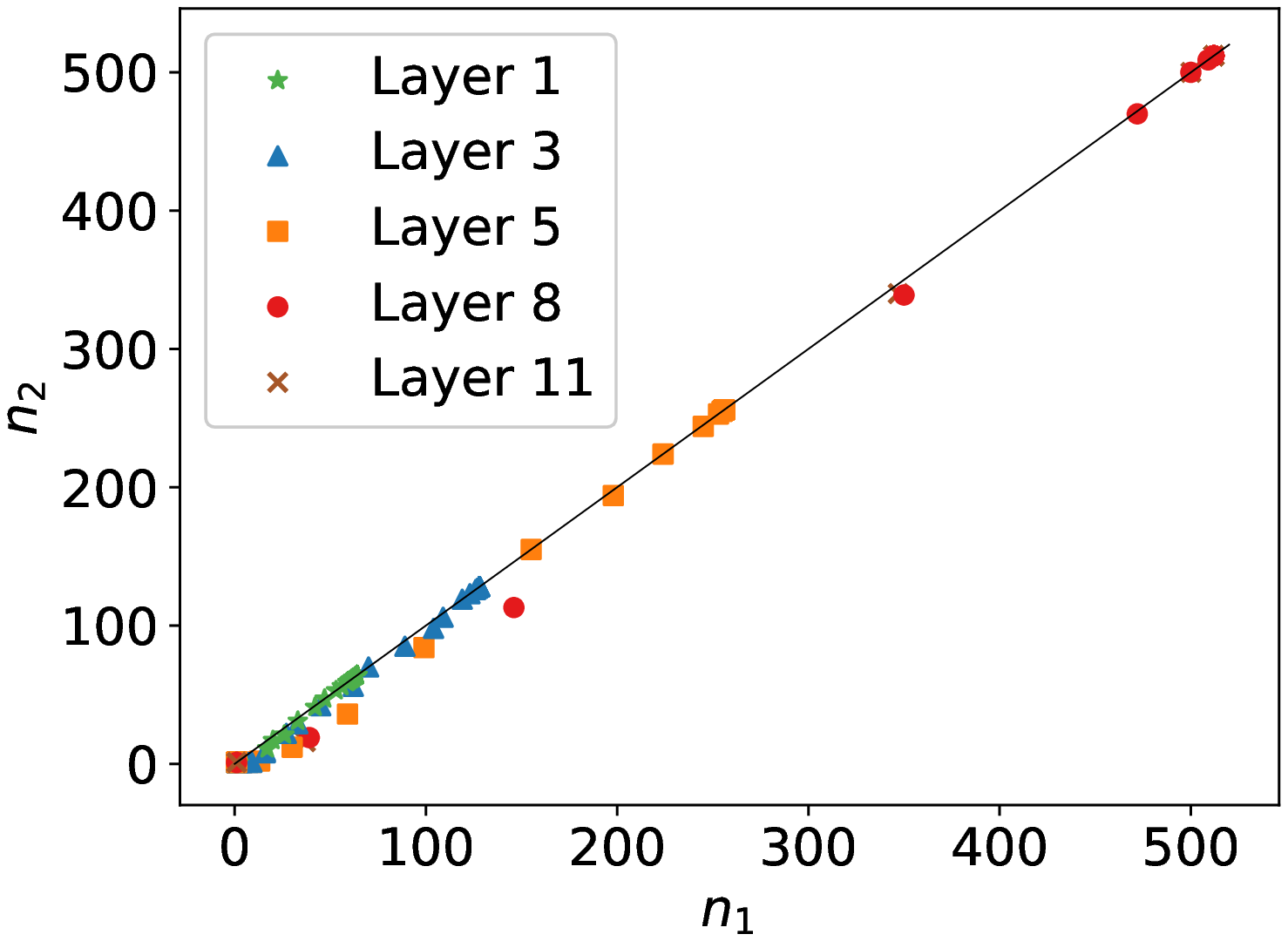}}
\subfigure[\# of filter reserved]{
\includegraphics[width=0.23\columnwidth]{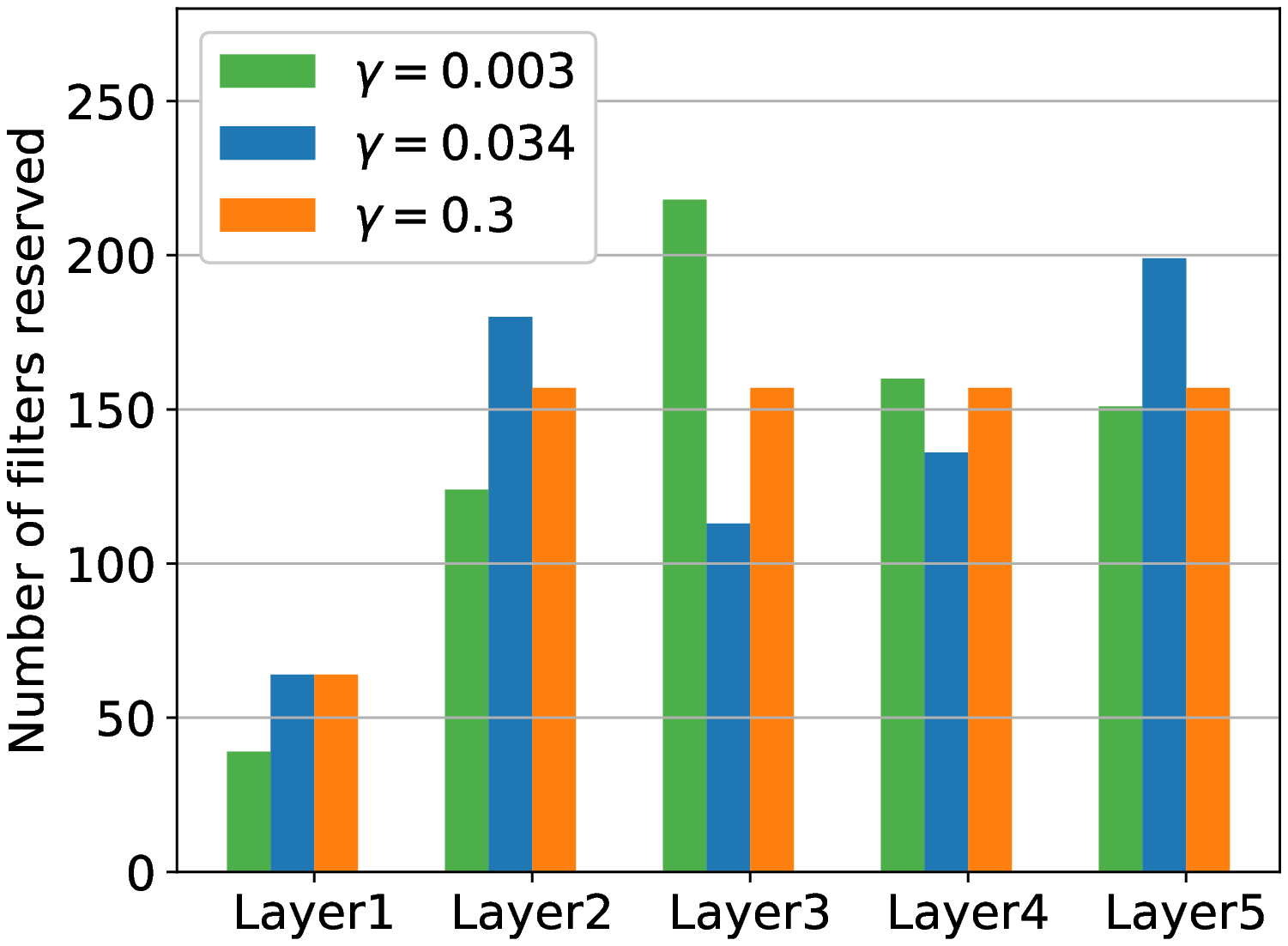}}
\subfigure[\% of filter reserved]{
\includegraphics[width=0.23\columnwidth]{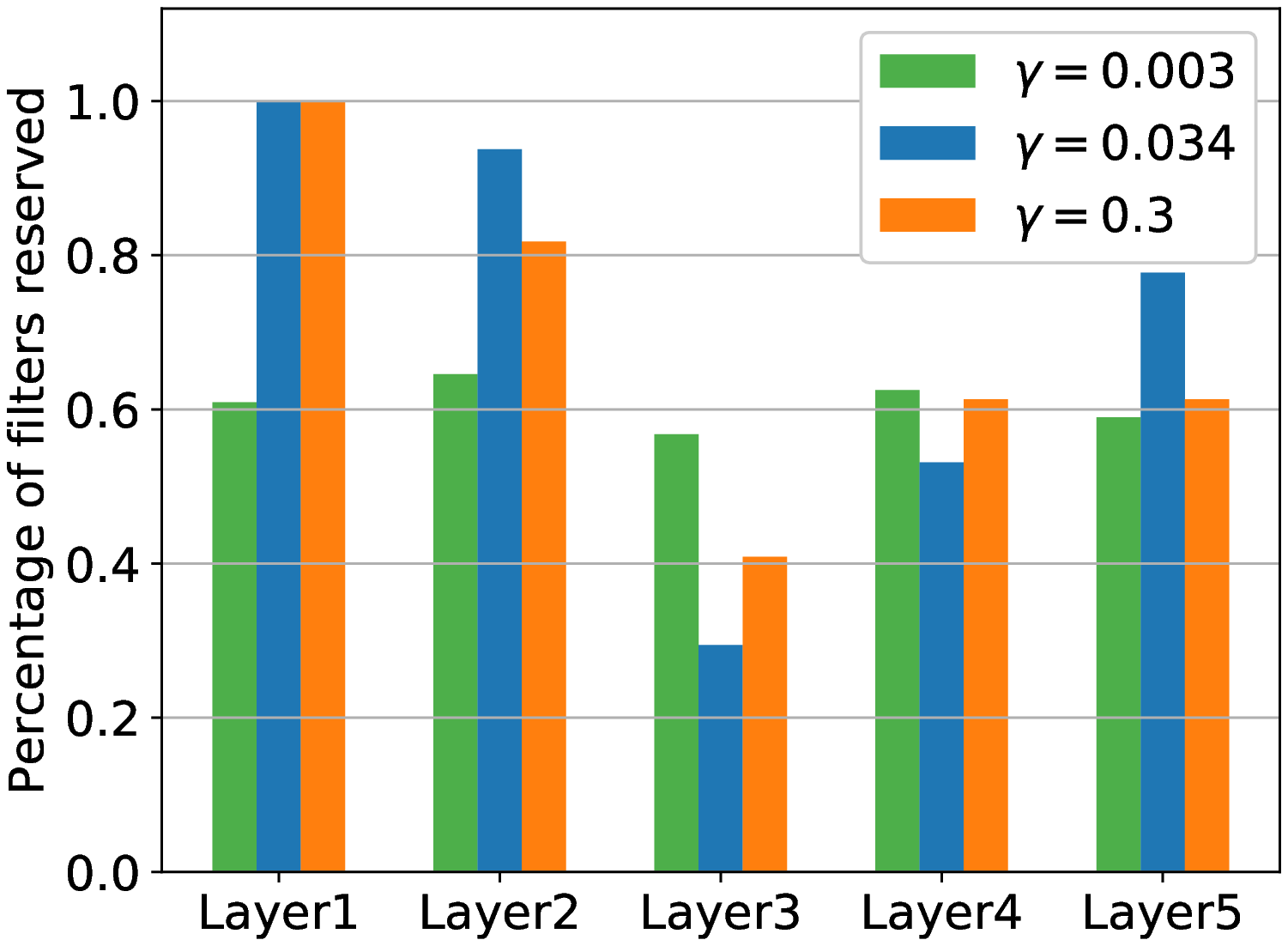}}
\end{minipage}
% \vskip -0.13in
\caption{(a-b) The value of $n_1$ and $n_2$ by changing $\gamma$ from 0.001 to 0.3. Black solid line refers to $n_1=n_2$. (c-d) The network structure comparison when $40\%$ filters are pruned from AlexNet using different $\gamma$s.}
\label{fig:12degree}
% \vskip -0.1in
\end{figure*}

We first show the effectiveness of our approach for l-covering number estimate. We build a series of graphs for each layer of a pre-trained AlexNet and VGG16 by changing $\gamma$ from $0.001$ to $0.3$ and visualize $n_1$ and $n_2$ (defined in the Methodology section) for the illustration purpose (Fig.~\ref{fig:12degree}(a-b)). Obviously, $n_1 \approx n_2$ in nearly all cases. The same trend is also observed in ResNet. Actually, in all of our experiments, we do not observe any large deviations between $n_1$ and $n_2$. It is with negligible influence to use $\tilde{N_1^c}=\frac{1}{2}(n_1+n_2)$ as an estimate of $N_1^c$, whose values is between $n_1$ and $n_2$.
% \vskip -0.2in

We also evaluate the computing time for estimating the 1-covering number $N_1^c$. To obtain $N_1^c$, a complete search of all combinations of vertices has to be done to see if all the vertices are covered by the selected balls, We name this approach as the oracle approach. We measure the computation time of the oracle approach and our proposed lightweight approach for calculating 1-covering number. It is clear that with the oracle method, the computation time grows drastically as the number of filters and the actual value of $N_1^c$ increase. It is even not temporally feasible to use the oracle approach when $N_1^c>4$. In contrast, the time used with our method for $\tilde{N_1^c}$ is negligibly short and is merely influenced by the number of filters in the layer and the value of $\tilde{N_1^c}$. These results indicate that our proposed approach for the estimate of $N_1^c$ is valid and efficient. Therefore, the real running time of the proposed approach is almost the same as existing methods with the same filter selection criteria.

\begin{table}[t]
% \vskip -0.25in
\centering
% \vskip -0.1in
\begin{tabular}{|c|c|c|c|c|}
\hline
Approach & MW & MA & Taylor & Random \\
\hline
Accuracy & 75.99\% & 75.84\% & 75.95\% & 75.82\% \\
\hline
\end{tabular}
% \vskip -0.1in
\caption{Performance with different filter selection criteria after pruning 30\% FLOPs of AlexNet.}
\label{tab:criteria}
\vskip -0.1in
\end{table}

\begin{table*}[t]
\centering
\begin{tabular}{|C{1.555cm}|C{1.5cm}|C{1.5cm}|C{1.6cm}|C{1.6cm}|C{0.9cm}|C{0.9cm}|C{0.88cm}|}
\hline
\multirow{2}{*}{Approach} & Top-1 acc.  & Top-5 acc. & Top-1 acc. & Top-5 acc. & Top-1 & Top-5 & FLOPs  \\
& baseline & baseline & after prune & after prune & acc. $\downarrow$ & acc. $\downarrow$ & $\downarrow$ \\
% \hline
\Xhline{1.8\arrayrulewidth}
SRR-NOF & \multirow{3}{*}{76.13\%} & \multirow{3}{*}{92.86\%} & 74.88\% & 92.27\% & 1.25\% & 0.59\% & 44.0\%\\
% \hline
SRR-PCA &  & & 75.19\% & 92.48\% & 0.94\% & 0.38\% & 44.1\%\\
% \hline
SRR-GR & &  & 75.76\% & 92.67\% & 0.37\% & 0.19\% & 44.1\%\\
\hline
\end{tabular}
\caption{Performance of the pruned ResNet50 networks with different redundancy reduction measurements.}
\label{tab:rr_compare}
% \vskip -0.1in
\end{table*}

\subsection{Filter selection criteria}
In previous experiments, we use the minimum weight criterion to prune filters. We further investigate whether different filter selection criteria have any influence on the performance. We use AlexNet on CIFAR-10 for illustration, by pruning $30\%$ FLOPs and fine-tuning the remaining networks for $100$ epochs with a learning rate of $1e^{-4}$. Results (Table~\ref{tab:criteria}) indicate that choosing filters with the minimum weight strategy achieves the best accuracy ($75.99\%$). However, using other filter selection criteria results in a similar performance, and the accuracy only drops $0.17\%$ even we randomly prune filters in the layers identified as with large redundancy by our approach. Therefore, our proposed approach is not sensitive to filter selection criteria, which further validates the fact that pruning filters in the layers with large redundancy is more essential than identifying unimportant filters.

\subsection{The value of gamma}
We change the distance threshold $\gamma$ for graph establishment to analyze its influence on the performance. We keep using AlexNet on CIFAR-10 as an example by pruning $40\%$ of the filters with $\gamma=0.003,0.034$, and $0.3$. Results (Fig.~\ref{fig:12degree}(c-d)) shows that with a large $\gamma$, the last four layers remain the same number of filters, which indicates that the layer with the largest number of filters are identified as the redundant layer at each time (Fig.~\ref{fig:12degree}(c)). When $\gamma$ is small, nearly the same percent of filters are removed from all layers (Fig.~\ref{fig:12degree}(d)). With a suitable value of $\gamma$ ($0.034$), our approach identified Layer 3 as the most redundant layer. Layer 4 and 5 are also redundant to some extent but Layer 4 is with a little more redundancy. Different from other existing works, our approach suggests that Layer 1 should not be pruned if we only aim to remove $40\%$ of the filters from AlexNet. These results are consistent with the definition of layer redundancy (in the Methodology section). For a layer with $n$ filters, when $\gamma \to 0$, it is clear that $k=n$, $N_1^c=n$, and $R(X)=1$. Therefore, all layers have the same level of redundancy and the approach becomes a uniform pruning. When $\gamma \to +\infty$, $k=1$, $N_1^c=1$, and $R(X)=n$, which indicates that the layer with the most number of filters is with the largest redundancy. Our approach can be considered as a dynamic architecture search approach controlled by $\gamma$.
% \vskip -0.22in
% \begin{figure}[t]
% \begin{center}
% \centerline{
% \subfigure[\# of filter reserved]{
% \includegraphics[width=0.25\columnwidth]{figs/distribution_number_alex_prune40.eps}}
% \subfigure[\% of filter reserved]{
% \includegraphics[width=0.25\columnwidth]{figs/distribution_percent_alex_prune40.eps}}
% }
% \vskip -0.15in
% \caption{The network structure comparison when $40\%$ filters are pruned from AlexNet using different $\gamma$s.}
% \label{fig:distribution}
% \end{center}
% \vskip -0.36in
% \end{figure}

% value selection is still sensitive and need hand-craft, automatic gamma selection as future work?

% \subsection{Time used for redundancy calculation over number of filters}
% or over different architectures.

\subsection{Other criteria for structural redundancy identification}
Since there are few studies that consider network pruning from the perspective of structural redundancy reduction, we further investigate the effectiveness of structural redundancy reduction for channel pruning with the following layer redundancy measurement metrics. (1) SRR-NOF: This strategy simply uses the number of filters in the convolutional layers as the measurement of layer redundancy. The layer with more filters is considered as with more redundancy. In the redundancy identification phase, for each iteration, a filter from the layer with the most number of filters are removed. If there exist more than one layer containing the same number of the most filters, a filter is removed from a randomly chosen layer. When the requirement is reached, the network is pruned with the minimum weight criterion according to the remaining number of filters in each layer. (2) SRR-PCA: This strategy uses principal component analysis (PCA) on the intermediate feature maps of a network to measure the correlation between filters. We first feed training samples to the CNN and record the flattened feature maps of each convolutional layer. Then we fit PCA on these flattened feature maps and get a list of percentage of variance explained by each of them for all convolutional layers. We select the $N$ smallest percentage of variances across all layers and count how many items are selected in each convolutional layer. Finally, we prune the filters in each layer accordingly, with the minimum weight ranking criterion. (3) SRR-GR: This strategy uses graph redundancy as described in the Methodology section. The training and pruning configuration are the same as in the Experiment section. 

Experiment results (Table \ref{tab:rr_compare}) show that by pruning $44\%$ FLOPs from ResNet50, even with a naive layer redundancy measurement (i.e., the number of filters in the layer), the performance is comparable to recent studies. PCA identifies the layer redundancy better than NOF and with SRR-PCA, the drops of top-1 and top-5 accuracy further decrease to $0.94\%$ and $0.38\%$. With the graph redundancy-based approach, the pruning performance is significantly improved. These results validate that (1) structural redundancy reduction is an efficient approach for channel pruning, and (2) the proposed graph redundancy-based approach is a promising way for layer redundancy measurement.

\section{Conclusion}
\label{sec:con}
We theoretically studied the rationale behind network pruning from a perspective of redundancy reduction via a statistical modeling and discovered that pruning filters in the layer(s) with the most structural redundancy plays a more essential role than pruning the least important filters across all layers. We proposed to identify the level of redundancy existed in each convolutional layer of a CNN via a graph establishment for each layer and two graph-related quantities as the measurement of the redundancy. After that, filters are pruned from the selected layer(s) by a simple filter selection criterion. Experimental results validated that our approach improved the state-of-the-art on image classification tasks. We believe that the proposed approach can be effective on more complicated tasks such as object detection and image synthesis, which is left for future research.

{\small
\bibliographystyle{ieee_fullname}
\bibliography{egbib}
}

\end{document}